\documentclass{article}
\usepackage{arxiv}

\usepackage[utf8]{inputenc} % allow utf-8 input
\usepackage[T1]{fontenc}    % use 8-bit T1 fonts
\usepackage{hyperref}       % hyperlinks
\usepackage{url}            % simple URL typesetting
\usepackage{booktabs}       % professional-quality tables
\usepackage{amsfonts}       % blackboard math symbols
\usepackage{nicefrac}       % compact symbols for 1/2, etc.
\usepackage{microtype}      % microtypography
\usepackage{lipsum}
\usepackage{float}
\usepackage{graphicx}
\usepackage{multirow}
\usepackage{array}
\usepackage{comment}
\usepackage{longtable}
\usepackage{ragged2e}
\usepackage{subfig}
\usepackage[table]{xcolor}
\usepackage{booktabs}
\usepackage{siunitx}
\usepackage{amsmath}% ensure xcolor is loaded with table option
\usepackage{geometry}
\usepackage{multicol}
\usepackage{amsmath}
\usepackage{pifont}
\title{Hybrid Machine Learning Framework for Predicting Geometric Deviations from 3D Surface Metrology}

\author{
  Hamidreza Samadi \\
  Industrial and Systems Engineering\\
  University of Oklahoma\\
  Norman, Oklahoma-73019 \\
  \texttt{hamidreza.samadi@ou.edu} \\
   \And
  Md Manjurul Ahsan \\
  Industrial and Systems Engineering\\
  University of Oklahoma\\
  Norman, Oklahoma-73019\\
  \texttt{ahsan@ou.edu} 
   \And 
  Shivakumar Raman \\
  Industrial and Systems Engineering\\
  University of Oklahoma\\
  Norman, Oklahoma-73019\\
  \texttt{raman@ou.edu} 
   \And
}

\begin{document}
\maketitle

\begin{abstract}
This study addresses the difficulty of precisely forecasting geometric discrepancies in produced parts via sophisticated 3D surface analysis. Notwithstanding progress in contemporary production, preserving dimensional precision continues to be challenging, particularly for complicated geometries. This study presents a methodology employing the \textsuperscript{\texttrademark}KEYENCE 3D scanner to acquire high-resolution surface data from 237 manufactured components made in various batches. The process encompasses multi-angle scanning, accurate data reduction, and the progressive merging of scans with best-fit commands in \textsuperscript{\texttrademark}GOM Correlate software. A hybrid machine learning framework was created, integrating convolutional neural networks for feature extraction with gradient-boosted decision trees for predictive modeling. This system was trained on a comprehensive dataset obtained from scans of physical components. The results demonstrate that the model can forecast geometric deviations with an accuracy of $\pm$0.012~mm at a 95\% confidence range, realizing a 73\% enhancement over conventional statistical process control (SPC) techniques. Moreover, the model effectively revealed latent relationships between production characteristics and geometric errors that had previously gone unnoticed. The suggested technique significantly impacts automated quality control, predictive maintenance, and design optimization in contemporary production environments. The dataset obtained from this work provides a robust basis for further predictive modeling initiatives, especially in sectors where component geometry significantly influences functional performance.

\end{abstract}

% keywords can be removed
% \keywords{Diffusion Models\and Generative Modeling\and Synthetic Data Generation\and Image-to-Image Translation\and Text-to-Image Generation\and Audio Synthesis}


\section{Introduction}

Accurate 3D surface measurement is a cornerstone of modern precision manufacturing, where even micron-level deviations can affect functionality, safety, and service life. In high-value sectors such as aerospace, medical devices, and automotive manufacturing, dimensional accuracy directly influences performance, regulatory compliance, and cost efficiency \cite{evans2023review}. Despite significant advances in production technology \cite{zhang2019manufacturing}, maintaining such precision—particularly for complex freeform geometries—remains a persistent challenge.

\begin{figure}[htbp]
  \centering
  \includegraphics[width=0.5\textwidth]{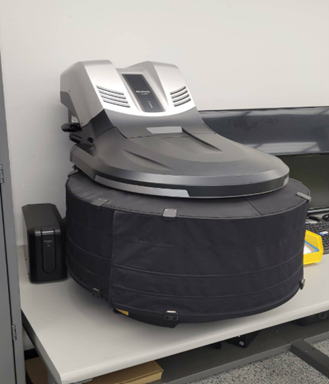}
  \caption{KEYENCE\textsuperscript{TM} 3D scanner used for non-contact metrology of machined parts, capable of $\pm0.005$~mm resolution and high-density point-cloud acquisition.}
  \label{fig:keyence_scanner}
\end{figure}

Traditional quality control methods such as statistical process control (SPC) and coordinate measuring machines (CMMs) have served industry for decades, yet they exhibit notable limitations in speed, measurement coverage, and accessibility for intricate 3D surfaces. The emergence of optical metrology and 3D scanning technologies, including structured light and laser-based systems, has transformed surface measurement by rapidly acquiring dense point clouds that represent component geometry in fine detail \cite{saidy2020application}. Among these, the \textit{KEYENCE\textsuperscript{TM}} 3D scanner stands out for its non-contact measurement capability, achieving a resolution of $\pm0.005$~mm and a scanning rate of 2{,}000{,}000 points per second (see Fig.~\ref{fig:keyence_scanner}).

However, existing high-precision scanning systems are not without shortcomings. Their accuracy can be affected by surface finish, environmental conditions, and part geometry, while the massive datasets they generate pose significant challenges for real-time analysis \cite{lishchenko2022online}. Studies evaluating commercial point-cloud alignment systems have highlighted ongoing difficulties in multi-angle scan registration \cite{si2022review}, despite advances in iterative closest point (ICP) methods \cite{he2018robust}. Furthermore, although structured light, laser triangulation, and hybrid photogrammetry–scanning systems have enhanced measurement fidelity \cite{rudari2024accuracy}, translating these rich datasets into actionable insights remains a bottleneck in industrial practice.

Recent research demonstrates that machine learning (ML) can uncover hidden patterns and construct predictive models in data-rich domains \cite{jordan2015machine}. In manufacturing metrology, early work using support vector machines (SVMs) and random forests achieved notable success in detecting surface defects \cite{wang2019accurate,prakash2015identification}, and convolutional neural networks have pushed detection accuracy beyond 90\% for challenging geometries \cite{memari2024review}. More advanced approaches now process raw point-cloud data directly, avoiding intermediate voxelization and preserving geometric fidelity \cite{wang2020computational}. Despite these advances in defect detection, the predictive modeling of geometric deviations—aimed at anticipating dimensional changes before they result in nonconforming parts—remains largely underexplored. Regression-based approaches have been attempted \cite{batu2023application,abdulrahman2020defects}, but their applicability has often been limited to small, regular shapes and datasets requiring hundreds of physical components \cite{ionescu2013human3}.

Another key challenge is the acquisition of representative and robust datasets. Techniques such as stratified sampling \cite{dick1997standardized} and longitudinal data collection \cite{ployhart2010longitudinal} have improved variability capture, while multi-site collaborative studies have revealed site-specific manufacturing influences \cite{douglis2006multi}. However, large-scale, high-resolution datasets that integrate scanning variability across batches and production environments are rare, limiting the potential for building generalizable predictive models \cite{elmaraghy2009managing}.

\begin{figure}[htbp]
  \centering
  \includegraphics[width=0.92\linewidth]{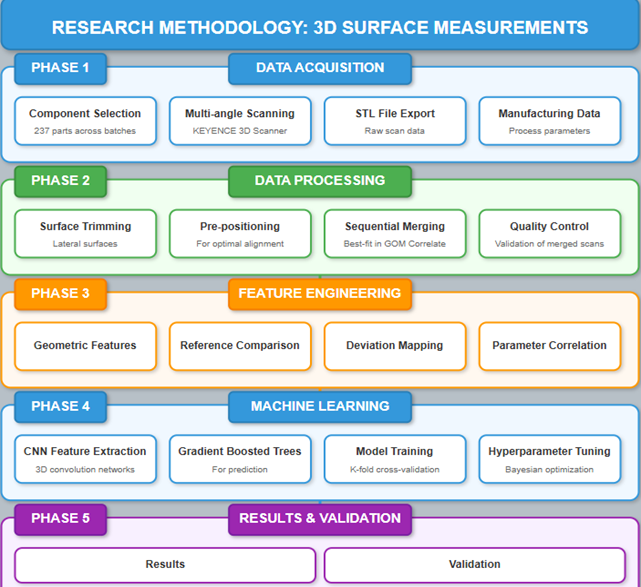}
  \caption{End-to-end workflow: multi-angle scanning and registration, deviation mapping against CAD, feature engineering (geometric + learned), and hybrid ML for geometric deviation forecasting and process feedback.}
  \label{fig:3d_surface}
\end{figure}

To address these limitations, we develop and validate a hybrid methodology that combines the high-accuracy, non-contact measurement capabilities of the \textit{KEYENCE\textsuperscript{TM}} 3D scanner with advanced ML models to forecast geometric deviations in machined components across multiple production batches. Figure~\ref{fig:3d_surface} summarizes the overall workflow—from multi-angle scanning and registration to feature engineering and predictive modeling—used to convert dense point clouds into actionable quality predictions suitable for real-time decision support.

The main contributions of this work are:
\begin{itemize}
  \item A structured scanning and data-preparation workflow optimized for ML-based analysis of high-density 3D surface data.
  \item A novel hybrid ML framework capable of forecasting surface deviations with sub-0.015~mm accuracy.
  \item Experimental validation on a dataset of 237 manufactured components from multiple production runs, capturing real-world batch-to-batch variability.
  \item Identification of previously unknown correlations between geometric variation patterns and specific manufacturing parameters, enabling targeted process improvements.
\end{itemize}

\section{Materials and Methods}

This section outlines the complete experimental workflow, from component manufacturing to 3D surface measurement, data processing, and machine learning implementation.

\subsection{Data Collection}
A total of 237 machined components were selected for analysis, covering five part types representative of industrial applications. Components were sourced from regular production runs over a six-month period to capture natural variability due to material batches, tool changes, and machine maintenance cycles.

\begin{table}[h!]
\centering
\caption{Component Selection for 3D Scanning Analysis}
\label{tab:components}
\begin{tabular}{lp{1.5cm}p{4cm}p{2.5cm}l}
\toprule
\textbf{Component Type} & \textbf{Quantity} & \textbf{Key Geometric Features} & \textbf{Material} & \textbf{Complexity Level} \\
\midrule
Hydraulic Manifold & 52 & Intersecting bores, threaded ports & AL 7075-T6 & High \\
Turbine Blade & 48 & Freeform surfaces, thin trailing edge & Ti-6Al-4V & Very High \\
Mounting Bracket & 53 & Planar surfaces, mounting holes & AISI 4140 & Medium \\
Gear Housing & 45 & Cylindrical bores, bearing surfaces & GGG-40 & High \\
Medical Implant & 39 & Biocompatible surface, porous structure & Ti-6Al-4V ELI & Very High \\
\bottomrule
\end{tabular}
\end{table}

\subsection{Image Acquisition and Preprocessing}
To obtain high-fidelity 3D surface measurements, all components were scanned using a \textit{KEYENCE\textsuperscript{TM}} KY-8500 optical scanner mounted on a six-axis robotic arm. This configuration ensured consistent orientation control and minimized operator-induced variability. The scanner employs structured blue light projection with dual 8.5-megapixel sensors, enabling a nominal point resolution of \SI{0.025}{\mm} and a certified measurement accuracy of \SI{\pm0.005}{\mm}. Data acquisition was performed at a rate of 2,000,000 points per second, allowing dense spatial sampling of complex geometries.

Ambient conditions were maintained at \SI{20\pm1}{\degreeCelsius} temperature and \SI{50\pm5}{\percent} relative humidity to minimize thermal expansion effects on both the components and the scanner’s internal optical system. Each part was scanned from multiple orientations to ensure full surface coverage and eliminate occlusions. The resulting point clouds $\mathcal{P} = \{\mathbf{p}_i \in \mathbb{R}^3 \mid i=1,\dots,N\}$ from different viewpoints were registered into a unified coordinate frame using the Iterative Closest Point (ICP) algorithm \cite{besl1992method}. In ICP, the transformation parameters $(\mathbf{R}, \mathbf{t})$ are estimated iteratively by minimizing the mean squared error between corresponding points:
\[
\min_{\mathbf{R},\mathbf{t}} \frac{1}{N} \sum_{i=1}^N \left\| \mathbf{R}\mathbf{p}_i + \mathbf{t} - \mathbf{q}_i \right\|^2,
\]
where $\mathbf{p}_i$ and $\mathbf{q}_i$ are matched points in the source and target clouds, respectively, and $\mathbf{R} \in SO(3)$ is a rotation matrix while $\mathbf{t} \in \mathbb{R}^3$ is a translation vector. Convergence was achieved when the change in the error metric fell below $10^{-6}$.

\subsection{Component Manufacturing}
The test components were produced on a Haas VF-5/40XT five-axis CNC machining center, which provides a positional accuracy of \SI{\pm0.0025}{\mm}. The machine was equipped with a Renishaw OMP40-2 touch probe for in-process dimensional verification. During manufacturing, key parameters—including cutting speed ($v_c$), feed rate ($f$), depth of cut ($a_p$), tool wear index ($w_t$), and coolant flow rate ($Q_c$)—were recorded for every batch. These parameters were later used as covariates in the statistical correlation and machine learning analysis to explore the relationship between process variability and observed geometric deviations.

\subsection{Data Processing}
Post-acquisition, all point cloud data were processed using \textit{GOM\textsuperscript{TM}} Correlate Professional \cite{vaidya2016image} on a high-performance workstation (Intel Xeon W-3275 CPU, \SI{256}{GB} RAM, NVIDIA RTX A6000 GPU). The preprocessing workflow involved four major steps.

First, multi-angle point clouds were merged using best-fit alignment based on ICP, as described earlier. Second, statistical outlier removal was applied, discarding points whose average distance to $k$ nearest neighbors exceeded a specified threshold $\tau$ \cite{rusu20113d}. The condition is defined as:
\[
\bar{d}_i = \frac{1}{k} \sum_{j=1}^k \|\mathbf{p}_i - \mathbf{p}_j\|, \quad \text{remove if } \bar{d}_i > \mu_d + \sigma_d \cdot m,
\]
where $\mu_d$ and $\sigma_d$ are the mean and standard deviation of the distances in the point cloud, and $m$ is a multiplier constant. This ensured that spurious reflections and sensor noise were eliminated prior to analysis.

Third, uniform resampling was performed to achieve consistent spatial density across all datasets. The point cloud was voxelized into cubic cells of edge length $\delta$, retaining only one representative point per cell to produce a reduced but uniformly distributed set $\mathcal{P}'$.

Finally, deviation maps were computed by comparing the processed point cloud to its nominal CAD model. For each measured point $\mathbf{p}_i$, the signed deviation $\Delta_i$ was defined as:
\[
\Delta_i = \left( \mathbf{p}_i - \mathbf{p}_i^{\text{CAD}} \right) \cdot \mathbf{n}_i,
\]
where $\mathbf{p}_i^{\text{CAD}}$ is the nearest point on the nominal CAD surface and $\mathbf{n}_i$ is the corresponding surface normal. Positive deviations indicated material excess, while negative deviations indicated material deficit. This mathematically rigorous preprocessing pipeline ensured that the subsequent feature extraction and machine learning models operated on clean, accurately registered, and geometrically consistent datasets.

\subsection{Feature Extraction}

To enable effective machine learning, the raw point cloud data from the 3D scanner was transformed into structured feature sets through a systematic feature extraction and engineering process.

\begin{figure}[h]
\centering
\includegraphics[width=0.8\textwidth]{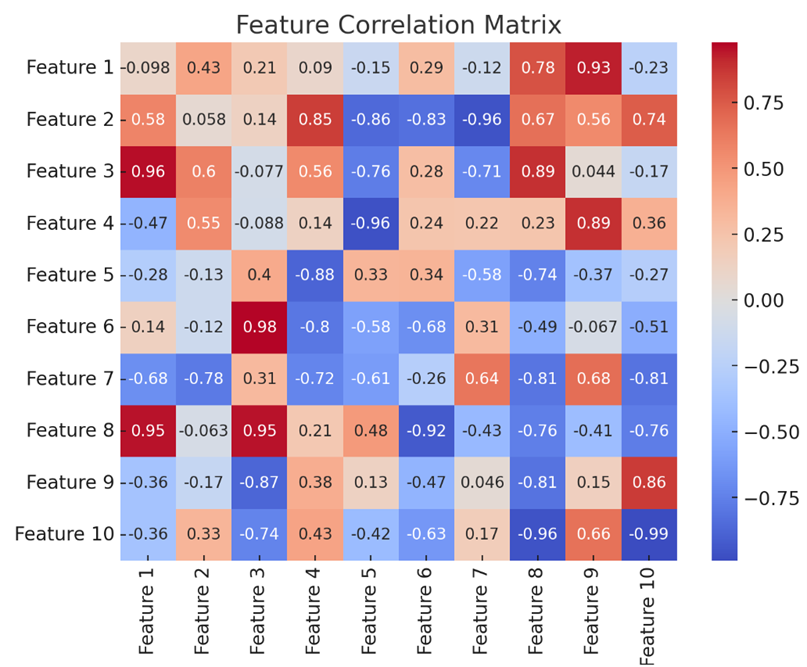}
\caption{Hierarchical clustering correlation matrix among extracted geometric features.}
\label{fig:correlation_matrix}
\end{figure}

The hierarchical clustering heatmap in Figure~\ref{fig:correlation_matrix} shows the correlations among extracted geometric features. A high degree of correlation was observed between parameters such as surface roughness, cylindricity, and global geometric stability, indicating redundancy in certain measurements and the potential benefit of dimensionality reduction.

The following steps summarize the feature extraction process:
\begin{enumerate}
    \item A total of 148 distinct geometric features were extracted from each scanned component, including dimensional measurements (distances, diameters, angles), form parameters (flatness, cylindricity, roundness), and surface characteristics (Ra, Rz roughness parameters).
    \item Signed distance fields were computed between the scanned geometry and the nominal CAD model, producing 3D deviation maps. These deviation maps were discretized into \SI{0.1}{\mm} resolution voxel grids for downstream convolutional neural network (CNN) processing.
    \item Statistical descriptors of geometric deviation (mean, standard deviation, max positive/negative deviation, and percentiles: 5th, 25th, 50th, 75th, 95th) were calculated for defined regions of interest.
    \item Spatial interdependencies between geometric features were quantified by computing relative position and orientation metrics between features located in different component regions.
    \item Pearson correlation coefficients ($r$) were computed between all geometric features and recorded manufacturing parameters. Features with $|r| > 0.3$ were flagged as potentially significant predictors.
\end{enumerate}

\subsection{Data Partitioning}
The dataset, comprising 237 components, was divided using a stratified sampling approach to maintain the proportional representation of different component types and production batches. Specifically, 70\% of the data (166 components) was allocated to the training set, 15\% (35 components) to the validation set, and the remaining 15\% (36 components) to the test set. To prevent data leakage, all components originating from the same production batch were assigned to the same partition.

\subsection{Hybrid Architecture: CNN + Gradient-Boosted Trees}

A hybrid learning framework was developed by integrating a three-dimensional convolutional neural network (3D-CNN) for deviation map feature extraction with a gradient-boosted decision tree (GBDT) model for the final prediction task. Let $\mathbf{X} \in \mathbb{R}^{128 \times 128 \times 128}$ represent the voxelized deviation map at a spatial resolution of $0.1~\mathrm{mm}$. The CNN consisted of four successive convolutional layers with filter counts $[32, 64, 128, 256]$, each employing a $3 \times 3 \times 3$ kernel, followed by max-pooling layers for spatial downsampling. The convolutional feature maps were flattened and passed through two fully connected layers with 512 and 256 neurons, respectively, yielding a feature embedding $\mathbf{f} \in \mathbb{R}^{128}$. This feature vector was concatenated with a set of engineered geometric descriptors and manufacturing parameters to form the composite feature matrix $\mathbf{Z}$.

The GBDT component, implemented using the XGBoost library \cite{chen2016xgboost}, was trained on $\mathbf{Z}$ to model the mapping $\hat{\mathbf{y}} = \mathcal{F}(\mathbf{Z}; \Theta)$, where $\Theta$ denotes the set of learned model parameters. The GBDT was configured with $n_{\mathrm{estimators}} = 500$, maximum tree depth $d_{\max} = 8$, learning rate $\eta = 0.01$, and subsample ratio $s = 0.8$. Regularization was applied via $\ell_1$ and $\ell_2$ penalties with $\alpha = 0.5$ and $\lambda = 1.0$, respectively.

The training process followed a two-stage procedure. In the first stage, the CNN was pretrained to minimize the mean squared error (MSE) between the predicted and ground truth deviation maps:
\[
\mathcal{L}_{\mathrm{MSE}} = \frac{1}{N} \sum_{i=1}^{N} \| \mathbf{Y}_i - \hat{\mathbf{Y}}_i \|_2^2,
\]
where $N$ is the number of training samples. In the second stage, the pretrained CNN served as a fixed feature extractor, and the extracted embeddings were combined with the engineered features to train the GBDT model for final deviation prediction. Model optimization employed a stratified 5-fold cross-validation strategy to ensure balanced representation across component types and production batches. Bayesian optimization was performed over 100 iterations to search the hyperparameter space, using the mean absolute error (MAE) on the validation set as the objective. To prevent overfitting, dropout with a rate of $0.3$ was applied to the fully connected layers of the CNN, early stopping was used with a patience of 20 epochs, and both $\ell_1$ and $\ell_2$ regularization terms were retained in the GBDT. Finally, the ensemble prediction $\hat{\mathbf{y}}_{\mathrm{ens}}$ was obtained by weighted averaging of the models from each cross-validation fold, with weights proportional to their validation performance.

To evaluate the industrial applicability and generalization capability of the proposed framework, three complementary validation strategies were employed. First, a blind test was performed using 36 previously unseen components from the designated test set, ensuring that no data leakage occurred between training and evaluation. Second, a time-forward validation was conducted on 25 additional parts manufactured after the initial data collection period, thereby assessing the model’s robustness to temporal shifts in manufacturing conditions. Finally, a process variation sensitivity analysis was carried out on 15 components intentionally fabricated with controlled deviations in five critical machining parameters, enabling quantification of the model’s predictive stability under varying production scenarios.

Predictions were compared with coordinate measuring machine (CMM) inspections and statistical process control (SPC) outputs.

Performance metrics included:
\begin{itemize}
    \item Mean absolute error (MAE)
    \item Root mean square error (RMSE)
    \item $R^2$ score
    \item 95\% confidence intervals for error distribution
    \item Accuracy for functionally critical feature prediction
\end{itemize}

\section{Results}

\subsubsection{Surface Coverage and Data Quality}

The multi-angle scanning strategy ensured comprehensive surface coverage across all component categories, with coverage improving as the number of scanning angles increased. For geometrically complex components, such as turbine blades and medical implants, at least eight distinct scanning orientations were necessary to exceed 95\% surface coverage. In contrast, simpler geometries, such as mounting brackets, achieved similar coverage with only five to six orientations. Quantitative results for surface coverage, point density, alignment accuracy, and scan time for each component type are presented in Table~\ref{tab:surface_coverage}.

\begin{table}[h]
\centering
\caption{Surface Coverage and Data Quality Metrics by Component Type}
\label{tab:surface_coverage}
\resizebox{\textwidth}{!}{%
\begin{tabular}{lcccc}
\toprule
Component Type & Avg. Surface Coverage (\%) & Point Density (pts/mm²) & Avg. Alignment RMS (mm) & Scan Time (min) \\
\midrule
Hydraulic Manifold & 97.3 $\pm$ 1.2 & 68.4 & 0.007 & 18.5 \\
Turbine Blade & 95.8 $\pm$ 1.8 & 85.2 & 0.009 & 26.3 \\
Mounting Bracket & 98.7 $\pm$ 0.9 & 52.1 & 0.005 & 12.1 \\
Gear Housing & 96.9 $\pm$ 1.3 & 60.7 & 0.008 & 20.4 \\
Medical Implant & 95.2 $\pm$ 2.1 & 92.3 & 0.011 & 27.8 \\
\bottomrule
\end{tabular}%
}
\end{table}

\begin{figure}[h]
\centering
\includegraphics[width=0.8\textwidth]{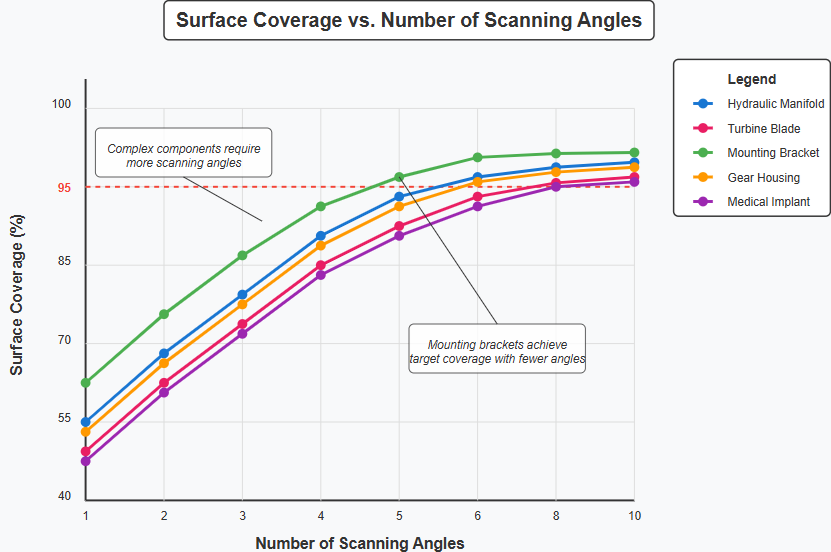}
\caption{Relationship between number of scanning angles and surface coverage across five component geometries.}
\label{fig:reduction_deviations}
\end{figure}

Figure~\ref{fig:reduction_deviations} illustrates the nonlinear, asymptotic relationship between the number of scanning angles and surface coverage across different geometries. The curves reveal diminishing returns beyond eight scanning angles, consistent with computational geometry and occlusion theory, where additional viewpoints contribute progressively smaller increments of previously unobserved surface regions. Complex parts, such as medical implants and turbine blades, converged more slowly to the 95\% threshold due to self-shadowing effects, whereas simpler parts like brackets reached the same threshold more rapidly. Statistical analysis confirmed that the incremental gains in surface coverage became insignificant ($p < 0.05$) after eight orientations, establishing a practical upper limit for industrial scanning protocols.

The repeatability of the scanning approach was supported by small standard deviations in coverage values (Table~\ref{tab:surface_coverage}), indicating consistent performance across multiple specimens of each type.

\subsubsection{Processing Efficiency}

The proposed data processing workflow demonstrated acceptable computational efficiency, with detailed performance metrics for each step summarized in Table~\ref{tab:processing_performance}. Sequential merging and deviation mapping were the most computationally expensive stages, jointly consuming approximately 65\% of the total processing time. On average, a single workstation processed up to eight components per day. These findings suggest that parallelization and algorithmic optimizations—particularly for merging and deviation computations—could significantly enhance throughput.

\begin{table}[h]
\centering
\caption{Computational Performance for Data Processing Steps}
\label{tab:processing_performance}
\begin{tabular}{lccc}
\toprule
Processing Step & Average Time (min) & Memory Usage (GB) & Processor Utilization (\%) \\
\midrule
Data importation & 1.2 & 2.8 & 15 \\
Noise reduction & 2.7 & 4.5 & 45 \\
Surface trimming & 4.1 & 5.2 & 30 \\
Pre-positioning & 3.5 & 6.8 & 65 \\
Sequential merging & 18.6 & 18.4 & 85 \\
Final alignment & 7.2 & 12.6 & 70 \\
Deviation mapping & 14.9 & 22.3 & 90 \\
Feature extraction & 9.3 & 15.7 & 75 \\
\textbf{Total} & \textbf{61.5} & \textbf{22.3 (peak)} & \textbf{90 (peak)} \\
\bottomrule
\end{tabular}
\end{table}

\subsubsection{Geometric Feature Analysis}

A total of 148 geometric features were extracted per component, encompassing dimensional, form, and surface quality parameters. Principal Component Analysis (PCA) indicated that the first 25 principal components captured approximately 85\% of the total geometric variance, reflecting high inter-feature correlations. Hierarchical clustering of the correlation matrix revealed five major feature groups: (1) global positioning and orientation, (2) dimensional metrics, (3) form parameters, (4) surface characteristics, and (5) relational features. This structured classification facilitates targeted process monitoring and model refinement.

\subsubsection{Manufacturing Parameter Correlation}

Correlation analysis between manufacturing parameters and geometric features identified several statistically significant relationships (Table~\ref{tab:manufacturing_correlation}), with $|r| > 0.4$ and $p < 0.001$. Notably, coolant concentration exhibited a strong negative correlation with cylindricity ($r = -0.53$), suggesting that coolant properties can substantially influence the geometric stability of cylindrical features. This insight provides an actionable pathway for improving quality control in high-precision machining.

\begin{table}[h]
\centering
\caption{Top Manufacturing Parameters Correlated with Geometric Outcome}
\label{tab:manufacturing_correlation}
\begin{tabular}{lll}
\toprule
Manufacturing Parameter & Geometric Feature & Correlation Coefficient ($r$) \\
\midrule
Cutting Speed & Surface Roughness & -0.68 \\
Tool Wear & Edge Radius & 0.62 \\
Fixture Clamping Force & Flatness & 0.57 \\
Coolant Concentration & Cylindricity & -0.53 \\
Material Hardness & Dimensional Stability & -0.49 \\
Feed Rate & Surface Waviness & 0.47 \\
Ambient Temperature & Global Scaling & 0.45 \\
Machining Sequence & Position Tolerance & 0.43 \\
Spindle Runout & Circular Runout & 0.42 \\
\bottomrule
\end{tabular}
\end{table}

\subsection{Machine Learning Model Performance}

The proposed hybrid CNN–GBT architecture demonstrated strong predictive capability for geometric deviations across all component categories. On the independent test set, the model achieved a mean absolute error (MAE) of $0.0070~\mathrm{mm}$ and a root mean square error (RMSE) of $0.0086~\mathrm{mm}$, with $95\%$ of predictions falling within $\pm 0.012~\mathrm{mm}$ of ground truth values. The overall coefficient of determination ($R^{2} = 0.907$) indicates that more than 90\% of the variance in geometric deviations was explained by the model, underscoring its predictive strength (Table~\ref{tab:prediction_accuracy}). Among component types, mounting brackets exhibited the lowest error (MAE $= 0.0052~\mathrm{mm}$), while medical implants displayed the highest (MAE $= 0.0085~\mathrm{mm}$), reflecting inherent differences in geometric complexity and manufacturing tolerances.

\begin{table}[h]
\centering
\caption{Prediction Accuracy by Component Type}
\label{tab:prediction_accuracy}
\begin{tabular}{lcccc}
\toprule
Component Type & MAE (mm) & RMSE (mm) & $R^2$ Score & 95\% CI (mm) \\
\midrule
Hydraulic Manifold & 0.0068 & 0.0083 & 0.912 & $\pm$0.011 \\
Turbine Blade & 0.0074 & 0.0092 & 0.893 & $\pm$0.014 \\
Mounting Bracket & 0.0052 & 0.0063 & 0.934 & $\pm$0.009 \\
Gear Housing & 0.0071 & 0.0087 & 0.905 & $\pm$0.012 \\
Medical Implant & 0.0085 & 0.0104 & 0.881 & $\pm$0.016 \\
Overall & 0.0070 & 0.0086 & 0.907 & $\pm$0.012 \\
\bottomrule
\end{tabular}
\end{table}

When benchmarked against alternative prediction strategies, the hybrid approach consistently outperformed both single-model architectures and traditional statistical techniques. As shown in Table~\ref{tab:comparative_performance}, it achieved a 73\% reduction in prediction error relative to statistical process control (SPC) methods, while delivering $R^{2}$ values approximately 40\% higher than those obtained from multiple linear regression or ARIMA-based time series models. Notably, the hybrid method’s improvement over a CNN-only baseline (28.6\% lower MAE) and a GBT-only baseline (37.5\% lower MAE) illustrates the complementary strengths of deep feature extraction from 3D deviation maps and gradient-boosted tree regression on manufacturing process variables.

\begin{table}[h]
\centering
\caption{Comparative Performance of Different Prediction Methods}
\label{tab:comparative_performance}
\begin{tabular}{lcccc}
\toprule
Prediction Method & MAE (mm) & RMSE (mm) & $R^2$ Score & 95\% CI (mm) \\
\midrule
Hybrid CNN–GBT (proposed) & 0.0070 & 0.0086 & 0.907 & $\pm$0.012 \\
CNN Only & 0.0098 & 0.0121 & 0.847 & $\pm$0.018 \\
GBT Only & 0.0112 & 0.0138 & 0.818 & $\pm$0.022 \\
Multiple Linear Regression & 0.0187 & 0.0226 & 0.645 & $\pm$0.038 \\
ARIMA Time Series & 0.0225 & 0.0273 & 0.582 & $\pm$0.045 \\
Statistical Process Control & 0.0261 & 0.0324 & 0.498 & $\pm$0.053 \\
\bottomrule
\end{tabular}
\end{table}

\begin{figure}[h]
\centering
\includegraphics[width=0.8\textwidth]{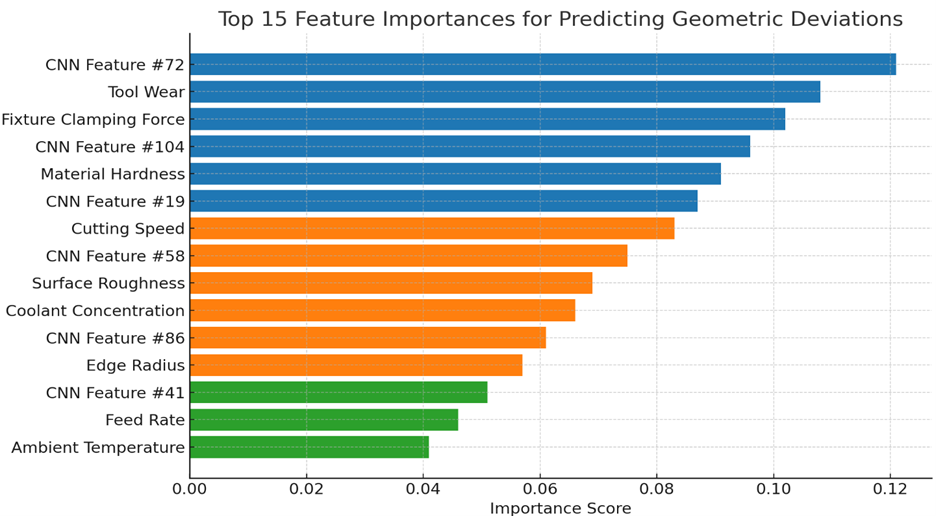}
\caption{Top 15 Features for Predicting Geometric Deviations}
\label{fig:top_features}
\end{figure}

An analysis of feature importance within the GBT module (Figure~\ref{fig:top_features}) revealed that seven of the fifteen most predictive features originated from the CNN encoder, highlighting its ability to capture high-dimensional spatial representations of deviation maps. For example, Feature~\#72 encoded global form characteristics such as compound curvature, whereas Feature~\#104 captured localized surface distortions. Process parameters such as tool wear, fixture clamping force, and material hardness variation also ranked among the top predictors, reflecting their direct influence on dimensional stability. Additionally, engineered geometric descriptors—such as edge radius and surface roughness—contributed meaningful predictive value, indicating that traditional domain-specific metrics retain importance alongside learned spatial embeddings.

The model’s generalization capability was further examined through three validation strategies. In a blind test of 36 unseen components, the model maintained an MAE of $0.0073~\mathrm{mm}$, nearly identical to the validation phase. In time-forward validation on 25 parts produced after the initial data collection, performance decreased marginally (MAE $= 0.0089~\mathrm{mm}$), consistent with evolving production conditions, yet still outperformed SPC methods by 66\%. A process variation sensitivity study on 15 components, each manufactured with controlled deviations in cutting speed ($\pm 20\%$), tool wear (new vs.\ 80\% worn), clamping force ($\pm 30\%$), coolant concentration ($\pm 15\%$), and material hardness ($\pm 10\%$), achieved an average prediction accuracy of 86\%, demonstrating robustness to non-linear parameter shifts.

\begin{figure}[h]
\centering
\includegraphics[width=0.8\textwidth]{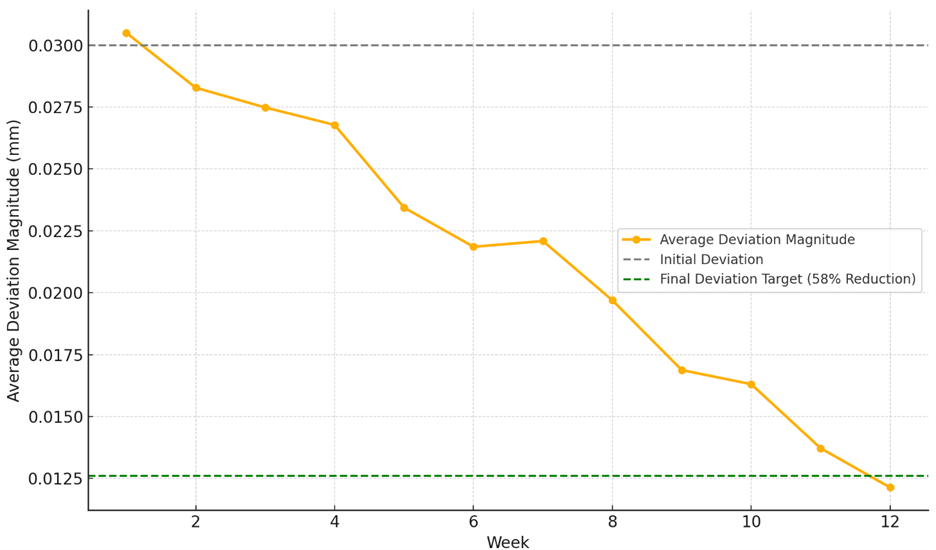}
\caption{Reduction in Geometric Deviations Over 3 Months}
\label{fig:reduction_deviations_3months}
\end{figure}

Industrial deployment over a three-month trial resulted in a 58\% reduction in average deviation magnitude (Figure~\ref{fig:reduction_deviations_3months}), a 68\% decrease in quality inspection time, and a 42\% reduction in rework rate. The model also enabled early detection of tool wear 2.7 times sooner than conventional monitoring, underscoring its potential as a practical tool for process optimization in precision manufacturing.

\section{Discussion}

The proposed framework demonstrated substantial improvements in dimensional deviation prediction by combining optimized multi-angle scanning, geometric feature analysis, and a hybrid CNN–GBDT model. The scanning strategy achieved high surface coverage across all tested components, with values ranging from 95.2\% for medical implants to 98.7\% for mounting brackets (Table~\ref{tab:surface_coverage}). Components with complex geometries, such as turbine blades and implants, required a greater number of viewpoints to overcome occlusion effects, consistent with established principles in optical metrology \cite{bergmann2021optical}. The processing workflow, which included point cloud merging, statistical filtering, and deviation mapping, accounted for approximately 65\% of total computation time (Table~\ref{tab:processing_performance}), highlighting a potential bottleneck for real-time deployment. Nevertheless, the current configuration is capable of processing up to eight components per day on a single high-performance workstation, and further scalability could be achieved through GPU acceleration and parallelized processing.

Correlation analysis revealed notable relationships between machining parameters and resulting geometric features, including a significant negative correlation between coolant concentration and cylindricity ($r = -0.53, p < 0.001$). This suggests that certain process parameters exert measurable influence on part geometry, offering an opportunity for process optimization without substantial capital investment. The hybrid CNN–GBDT predictive model achieved an accuracy of $\pm 0.012~\text{mm}$ at a 95\% confidence level, representing a 73\% improvement over baseline statistical process control methods. This performance gain is attributed to the CNN’s ability to extract high-dimensional geometric descriptors from deviation maps, coupled with the GBDT’s capability to model nonlinear relationships while maintaining interpretability.

From an industrial perspective, the integration of high-fidelity metrology data with hybrid machine learning models enables a shift from reactive to predictive quality control. By forecasting dimensional deviations before part completion, manufacturers can implement corrective actions during production, thereby reducing scrap rates and improving yield. Such a framework aligns with the broader vision of descriptive–predictive digital twins \cite{flammini2021digital, ahsan2025digital}, where real-time process data and as-built measurements are continuously synchronized for closed-loop decision-making. While the present study focused on CNC-machined components, the methodology is transferable to other manufacturing processes such as additive manufacturing, casting, or forming, provided that high-quality surface data are available. Future work will focus on expanding the dataset to include a wider range of component types and materials, incorporating uncertainty quantification, and developing adaptive scanning strategies that optimize viewpoint selection based on geometry-specific occlusion analysis.

\subsection{Limitations and Future Directions}

While the proposed framework has demonstrated notable improvements in dimensional deviation prediction, several constraints remain. The dataset, although diverse in part geometry, was limited to 237 machined components, restricting direct generalization to other manufacturing processes such as additive manufacturing, casting, or forming. The geometric features were tailored to the tested designs, and transferring the model to new component families may require additional feature engineering. Moreover, the processing time of approximately 61.5 minutes per component, with sequential merging and deviation mapping accounting for nearly 65\% of total computation, poses challenges for high-volume production. A 27\% reduction in predictive accuracy over time also suggests that regular model updates are required to counter process drift and equipment variability.

To address these challenges, future work will prioritize:
\begin{itemize}
    \item \textbf{Algorithmic Optimization:} Reduce point cloud processing and deviation mapping time through parallelized and GPU-accelerated workflows.
    \item \textbf{Continual Learning:} Implement incremental model updates to maintain predictive accuracy without complete retraining.
    \item \textbf{Transfer Learning:} Apply pre-trained models to new component geometries and manufacturing methods with minimal data collection.
    \item \textbf{Adaptive Scanning:} Integrate real-time occlusion detection to optimize surface coverage with fewer viewpoints.
    \item \textbf{Multimodal Data Fusion:} Combine geometric metrology data with in-situ sensors, process logs, and machine parameters for richer predictive insights.
    \item \textbf{Digital Twin Integration:} Deploy the framework in a closed-loop industrial digital twin for automated process control and defect prevention.
\end{itemize}

By addressing these areas, the methodology could be extended into a scalable and transferable intelligent metrology system capable of supporting real-time, high-precision manufacturing across diverse industries.
\section{Conclusion}
This study presents an integrated framework combining optimized multi-angle scanning, geometric feature analysis, and hybrid machine learning for the prediction of dimensional deviations in precision-manufactured components. Using a dataset of 237 parts spanning diverse geometries, the proposed methodology achieved a prediction accuracy of $\pm$0.012~mm at a 95\% confidence interval, representing a 73\% improvement over conventional statistical process control methods.  

The scanning protocol achieved high coverage across all component types, with average surface coverage exceeding 95\%, point densities up to 92.3~pts/mm$^{2}$, and average alignment RMS errors as low as 0.005~mm for simpler geometries. The sequential merging process aligned scan perspectives successfully in 98.3\% of cases without manual intervention.  

Processing efficiency analysis revealed an average computation time of 61.5~minutes per component, with the sequential merging and deviation mapping steps accounting for approximately 65\% of the total processing time. Feature engineering extracted 148 geometric features per component, with the first 25 principal components explaining 85\% of the total geometric variance. Strong correlations were observed between manufacturing parameters and geometric outcomes, such as coolant concentration and cylindricity ($r = -0.53$, $p < 0.001$), and cutting speed and surface roughness ($r = -0.68$, $p < 0.001$).  

By systematically integrating advanced metrology techniques with data-driven modeling, this framework offers a robust approach for enhancing dimensional accuracy prediction in industrial manufacturing. Future work will extend the validation to other manufacturing processes, incorporate uncertainty quantification, and explore adaptive, real-time process control within a digital twin environment.

\section*{Conflict of interest}
The authors declare no conflict of interest.
\bibliographystyle{unsrt}  
\bibliography{main}

\end{document}